\documentclass[runningheads]{llncs}
\usepackage[T1]{fontenc}
\usepackage{graphicx}
\usepackage{multirow}
\usepackage{algorithm}
\usepackage{amsmath}
\usepackage{amssymb}
\usepackage{booktabs}
\usepackage{hyperref}
\usepackage[misc]{ifsym}
\usepackage{hyperref}

\usepackage[table]{xcolor}
\usepackage{colortbl}

\definecolor{orange1}{RGB}{255,245,235} 
\definecolor{orange2}{RGB}{255,202,160}
\definecolor{orange3}{RGB}{255,179,120}
\definecolor{orange4}{RGB}{255,150,80}
\definecolor{orange5}{RGB}{255,120,40}
\definecolor{orange6}{RGB}{230,90,10}   

\newcommand{\colorcell}[2]{\cellcolor{#1}{#2}}
\usepackage{booktabs} 
\begin{document}
\title{How Far Have Medical Vision-Language Models Come? A Comprehensive Benchmarking Study}
\titlerunning{How Far Have Medical Vision-Language Models Come?}
%
\author{Che Liu\inst{1}\textsuperscript{(\Letter)} \and
Jiazhen Pan\inst{2} \and
Weixiang Shen\inst{3} \and \\ 
Wenjia Bai\inst{1} \and
Daniel Rueckert\inst{1,2}\and
Rossella Arcucci\inst{1}
}
%
\institute{Imperial College London, UK \and
Technical University of Munich, Germany \and
Ludwig Maximilian University of Munich
\\
\email{che.liu21@ic.ac.uk}}
\maketitle              
\begin{abstract}
Vision-Language Models (VLMs) trained on web-scale corpora excel at natural image tasks and are increasingly repurposed for healthcare; however, their competence in medical tasks remains underexplored. We present a comprehensive evaluation of open-source general-purpose and medically specialised VLMs, ranging from 3B to 72B parameters, across eight benchmarks: MedXpert, OmniMedVQA, PMC-VQA, PathVQA, MMMU, SLAKE, and VQA-RAD. To observe model performance across different aspects, we first separate it into understanding and reasoning components. Three salient findings emerge. First, large general-purpose models already match or surpass medical-specific counterparts on several benchmarks, demonstrating strong zero-shot transfer from natural to medical images. Second, reasoning performance is consistently lower than understanding, highlighting a critical barrier to safe decision support. Third, performance varies widely across benchmarks, reflecting differences in task design, annotation quality, and knowledge demands. No model yet reaches the reliability threshold for clinical deployment, underscoring the need for stronger multimodal alignment and more rigorous, fine-grained evaluation protocols.

\end{abstract}

\section{Introduction}
Vision Language Models (VLMs) have made remarkable progress in recent years, demonstrating strong capabilities in visual understanding, captioning, and multimodal reasoning when trained on large scale web datasets. General purpose VLMs such as \cite{bai2025qwen25vltechnicalreport,coreteam2025mimovltechnicalreport,pan2025medvlm} have shown impressive zero shot and few shot performance across a wide range of tasks, powered by advances in large language models and high capacity visual encoders. In contrast, the development and evaluation of VLMs in the medical domain remain relatively underexplored. This is due to several challenges unique to the medical setting, including limited access to paired image text data, a lack of large scale annotated resources, and the need for domain specific knowledge to interpret clinical language and visual content accurately. While several specialised medical VLMs (e.g., Lingshu \cite{xu2025lingshu}, Huatuo-GPT-Vision \cite{chen2024huatuogpt}) have been proposed, comprehensive and comparative evaluations remain limited.

\begin{figure}[h!]
    \centering
    \includegraphics[width=1\linewidth]{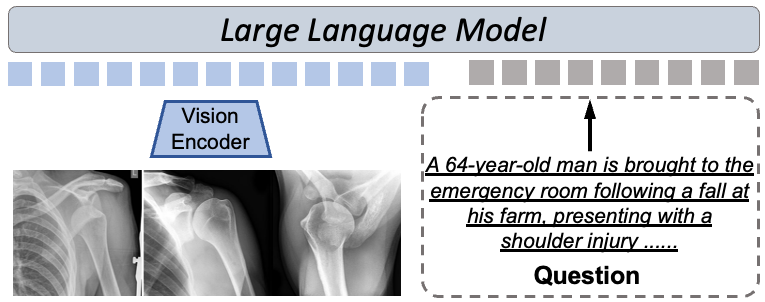}
    \caption{Illustration of a VLM processing medical images and a textual question.}
    \label{fig:vlm}
\end{figure}

At the same time, the community has introduced a number of medical vision language benchmarks covering tasks such as visual question answering, clinical report understanding, and multimodal reasoning. Datasets like MedXpert, OmniMedVQA, and VQA RAD provide valuable testing grounds, yet most existing studies focus on individual models or datasets in isolation. To address this gap, we present a comprehensive evaluation of both general purpose and medical specific VLMs on seven widely used medical benchmarks. To better understand model behaviour, we categorise tasks into two groups: understanding and reasoning. This enables a more fine grained analysis of model performance under different cognitive demands. We illustrate a standard VLM architecture in Figure~\ref{fig:vlm}.

\section{Benchmarking VLMs}
\subsection{Datasets and Task Coverage}

To evaluate medical VLMs comprehensively, we select seven widely used benchmarks that span diverse task types, imaging modalities, and annotation formats: \textbf{MedXpert}, \textbf{OmniMedVQA}, \textbf{PMC-VQA}, \textbf{PathVQA}, \textbf{MMMU} (medical subset), \textbf{SLAKE}, and \textbf{VQA-RAD}. These datasets collectively cover a broad spectrum of medical domains, including radiology, pathology, ophthalmology, dermatology, and ultrasound.

The benchmarks vary in their design: some focus on single image question answering (e.g., VQA RAD, SLAKE), while others involve multi image or multi view reasoning (e.g., MedXpert). The tasks encompass both visual understanding (e.g., identifying anatomical regions or imaging modalities) and reasoning that requires integrating medical knowledge and inferring from both visual and textual information.

\subsection{Decomposing Understanding and Reasoning}

Inspired by \cite{thapa2025disentangling}, we aim to disentangle the types of capabilities involved in medical VLM benchmarks by categorising each question into either \textit{understanding} or \textit{reasoning}. To do this, we train a classifier using samples from the MedXpert dataset, which provides explicit labels for these two categories. We use the pretrained \texttt{gme-Qwen2-VL-2B-Instruct}\footnote{\url{https://huggingface.co/Alibaba-NLP/gme-Qwen2-VL-2B-Instruct}} model to extract joint image–text embeddings for each sample, which are then used to train a binary classifier.
Once trained, the classifier is applied to all other datasets (except MedXpert) to infer whether each question primarily requires understanding or reasoning. This decomposition enables a fine-grained analysis of model performance across cognitive demands, independent of the dataset origin.

\subsection{Implementation Details}

We evaluate a diverse set of VLMs that includes both general-purpose and medically specialised models. The general-purpose group includes \texttt{Qwen2.5-VL}~\cite{bai2025qwen25vltechnicalreport} models at 3B, 7B, 32B, and 72B scales, as well as \texttt{MiMo-VL-7B}~\cite{coreteam2025mimovltechnicalreport} in both SFT and RL variants. The medical-specific group includes \texttt{Lingshu-7B}, \texttt{Lingshu-32B}~\cite{xu2025lingshu}, and \texttt{Huatuo-GPT-Vision-7B} and \texttt{32B}~\cite{chen2024huatuogpt}. All models are evaluated using their official inference codebases. For consistency across models, we use a standard prompt format: \texttt{“Please reason step by step, and put the final answer in \textbackslash boxed\{\}”}. We then extract the boxed answer and compare it against the ground truth option to compute accuracy.


\section{Experimental Results}
\begin{table}[t!]
\centering
\caption{Average performance of general-purpose and medical-specific VLMs across all benchmarks. Darker orange indicates higher per-column performance.}
\label{tab:vlm_performance}
\scalebox{0.8}{
\begin{tabular}{l|ccccccc}
\toprule[1.2pt]
Model & MedXpert & OmniMedVQA & PMC-VQA & PathVQA & MMMU & SLAKE & VQA-RAD \\
\midrule
\multicolumn{8}{l}{\textit{General-purpose VLMs}} \\
Qwen2.5-VL-3B              & \colorcell{orange1}{0.2217} & \colorcell{orange1}{0.5710} & \colorcell{orange1}{0.4014} & \colorcell{orange2}{0.6190} & \colorcell{orange1}{0.4690} & \colorcell{orange1}{0.5874} & \colorcell{orange1}{0.5458} \\
Qwen2.5-VL-7B              & \colorcell{orange1}{0.2305} & \colorcell{orange2}{0.6072} & \colorcell{orange2}{0.4420} & \colorcell{orange2}{0.6205} & \colorcell{orange2}{0.5379} & \colorcell{orange1}{0.6068} & \colorcell{orange4}{0.6255} \\
Qwen2.5-VL-32B             & \colorcell{orange5}{0.2890} & \colorcell{orange3}{0.6383} & \colorcell{orange5}{0.5331} & \colorcell{orange3}{0.6573} & \colorcell{orange6}{0.7075} & \colorcell{orange5}{0.7326} & \colorcell{orange6}{0.7041} \\
Qwen2.5-VL-72B             & \colorcell{orange6}{0.2995} & \colorcell{orange3}{0.6656} & \colorcell{orange6}{0.5577} & \colorcell{orange3}{0.6597} & \colorcell{orange6}{0.7078} & \colorcell{orange6}{0.7799} & \colorcell{orange6}{0.6835} \\
MiMo-VL-7B-RL              & \colorcell{orange2}{0.2446} & \colorcell{orange2}{0.6206} & \colorcell{orange2}{0.4585} & \colorcell{orange2}{0.6255} & \colorcell{orange4}{0.6000} & \colorcell{orange4}{0.7233} & \colorcell{orange3}{0.6135} \\
MiMo-VL-7B-SFT             & \colorcell{orange2}{0.2455} & \colorcell{orange2}{0.6323} & \colorcell{orange3}{0.4985} & \colorcell{orange2}{0.6398} & \colorcell{orange5}{0.6276} & \colorcell{orange4}{0.7257} & \colorcell{orange4}{0.6374} \\
\midrule
\multicolumn{8}{l}{\textit{Medical-specific VLMs}} \\
Lingshu-7B                 & \colorcell{orange3}{0.2505} & \colorcell{orange3}{0.6436} & \colorcell{orange4}{0.5213} & \colorcell{orange6}{0.7192} & \colorcell{orange5}{0.6389} & \colorcell{orange6}{0.8034} & \colorcell{orange5}{0.6574} \\
Lingshu-32B                & \colorcell{orange6}{0.3107} & \colorcell{orange6}{0.7662} & \colorcell{orange5}{0.5365} & \colorcell{orange6}{0.7609} & \colorcell{orange5}{0.6345} & \colorcell{orange6}{0.7718} & \colorcell{orange4}{0.6295} \\
Huatuo-GPT-Vision-7B       & \colorcell{orange1}{0.2215} & \colorcell{orange5}{0.7429} & \colorcell{orange6}{0.5490} & \colorcell{orange1}{0.5791} & \colorcell{orange2}{0.5448} & \colorcell{orange5}{0.7644} & \colorcell{orange2}{0.6056} \\
Huatuo-GPT-Vision-32B      & \colorcell{orange2}{0.2415} & \colorcell{orange6}{0.7619} & \colorcell{orange6}{0.5785} & \colorcell{orange2}{0.6291} & \colorcell{orange3}{0.5862} & \colorcell{orange6}{0.7764} & \colorcell{orange6}{0.6892} \\
\bottomrule[1.2pt]
\end{tabular}
}
\end{table}

\begin{table}[t!]
\centering
\caption{Performance of general-purpose and medical-specific VLMs on reasoning tasks across seven benchmarks. Darker orange indicates higher per-column performance.}
\label{tab:vlm_performance_reasoning}
\scalebox{0.8}{
\begin{tabular}{l|ccccccc}
\toprule[1.2pt]
Model & MedXpert & OmniMedVQA & PMC-VQA & PathVQA & MMMU & SLAKE & VQA-RAD \\
\midrule
\multicolumn{8}{l}{\textit{General-purpose VLMs}} \\
Qwen2.5-VL-3B              & \colorcell{orange1}{0.2151} & \colorcell{orange1}{0.5809} & \colorcell{orange1}{0.4251} & \colorcell{orange3}{0.5945} & \colorcell{orange1}{0.4865} & \colorcell{orange1}{0.5882} & \colorcell{orange1}{0.5340} \\
Qwen2.5-VL-7B              & \colorcell{orange2}{0.2254} & \colorcell{orange2}{0.6285} & \colorcell{orange2}{0.4738} & \colorcell{orange4}{0.6644} & \colorcell{orange2}{0.5495} & \colorcell{orange1}{0.6106} & \colorcell{orange4}{0.6456} \\
Qwen2.5-VL-32B             & \colorcell{orange5}{0.2718} & \colorcell{orange3}{0.6579} & \colorcell{orange5}{0.5507} & \colorcell{orange5}{0.6763} & \colorcell{orange6}{0.7170} & \colorcell{orange4}{0.7293} & \colorcell{orange6}{0.7487} \\
Qwen2.5-VL-72B             & \colorcell{orange6}{0.2953} & \colorcell{orange4}{0.6805} & \colorcell{orange5}{0.5757} & \colorcell{orange5}{0.6898} & \colorcell{orange6}{0.7264} & \colorcell{orange6}{0.7863} & \colorcell{orange6}{0.7236} \\
MiMo-VL-7B-RL              & \colorcell{orange2}{0.2350} & \colorcell{orange2}{0.6155} & \colorcell{orange3}{0.4886} & \colorcell{orange4}{0.6525} & \colorcell{orange3}{0.5946} & \colorcell{orange5}{0.7395} & \colorcell{orange3}{0.6262} \\
MiMo-VL-7B-SFT             & \colorcell{orange3}{0.2414} & \colorcell{orange2}{0.6284} & \colorcell{orange4}{0.5173} & \colorcell{orange5}{0.6746} & \colorcell{orange4}{0.6306} & \colorcell{orange4}{0.7311} & \colorcell{orange5}{0.6650} \\
\midrule
\multicolumn{8}{l}{\textit{Medical-specific VLMs}} \\
Lingshu-7B                 & \colorcell{orange3}{0.2427} & \colorcell{orange2}{0.6213} & \colorcell{orange4}{0.5380} & \colorcell{orange6}{0.7445} & \colorcell{orange5}{0.6455} & \colorcell{orange6}{0.8039} & \colorcell{orange5}{0.6748} \\
Lingshu-32B                & \colorcell{orange6}{0.2997} & \colorcell{orange6}{0.7526} & \colorcell{orange5}{0.5609} & \colorcell{orange6}{0.7462} & \colorcell{orange5}{0.6577} & \colorcell{orange6}{0.7815} & \colorcell{orange4}{0.6505} \\
Huatuo-GPT-Vision-7B       & \colorcell{orange1}{0.2033} & \colorcell{orange5}{0.7174} & \colorcell{orange5}{0.5720} & \colorcell{orange1}{0.5060} & \colorcell{orange2}{0.5225} & \colorcell{orange5}{0.7701} & \colorcell{orange2}{0.5720} \\
Huatuo-GPT-Vision-32B      & \colorcell{orange1}{0.2144} & \colorcell{orange6}{0.7473} & \colorcell{orange6}{0.6133} & \colorcell{orange3}{0.5894} & \colorcell{orange3}{0.5856} & \colorcell{orange6}{0.7922} & \colorcell{orange6}{0.7184} \\
\bottomrule[1.2pt]
\end{tabular}
}
\end{table}

\begin{table}[t!]
\centering
\caption{Performance of general-purpose and medical-specific VLMs on understanding tasks across seven benchmarks. Darker orange indicates higher per-column performance.}
\label{tab:vlm_performance_retrieval}
\scalebox{0.8}{
\begin{tabular}{l|ccccccc}
\toprule[1.2pt]
Model & MedXpert & OmniMedVQA & PMC-VQA & PathVQA & MMMU & SLAKE & VQA-RAD \\
\midrule
\multicolumn{8}{l}{\textit{General-purpose VLMs}} \\
Qwen2.5-VL-3B              & \colorcell{orange1}{0.1802} & \colorcell{orange1}{0.5500} & \colorcell{orange1}{0.3604} & \colorcell{orange1}{0.5851} & \colorcell{orange1}{0.4522} & \colorcell{orange1}{0.5552} & \colorcell{orange1}{0.5089} \\
Qwen2.5-VL-7B              & \colorcell{orange2}{0.2024} & \colorcell{orange2}{0.5801} & \colorcell{orange2}{0.3955} & \colorcell{orange2}{0.5984} & \colorcell{orange2}{0.4873} & \colorcell{orange1}{0.5706} & \colorcell{orange4}{0.5890} \\
Qwen2.5-VL-32B             & \colorcell{orange5}{0.2489} & \colorcell{orange3}{0.6210} & \colorcell{orange5}{0.4710} & \colorcell{orange4}{0.6403} & \colorcell{orange6}{0.5990} & \colorcell{orange5}{0.6841} & \colorcell{orange6}{0.6451} \\
Qwen2.5-VL-72B             & \colorcell{orange6}{0.2622} & \colorcell{orange4}{0.6395} & \colorcell{orange6}{0.5052} & \colorcell{orange5}{0.6611} & \colorcell{orange6}{0.6084} & \colorcell{orange6}{0.7013} & \colorcell{orange6}{0.6390} \\
MiMo-VL-7B-RL              & \colorcell{orange3}{0.2150} & \colorcell{orange2}{0.5999} & \colorcell{orange2}{0.4088} & \colorcell{orange2}{0.6040} & \colorcell{orange3}{0.5200} & \colorcell{orange5}{0.6722} & \colorcell{orange3}{0.5712} \\
MiMo-VL-7B-SFT             & \colorcell{orange3}{0.2257} & \colorcell{orange3}{0.6102} & \colorcell{orange3}{0.4399} & \colorcell{orange3}{0.6204} & \colorcell{orange4}{0.5622} & \colorcell{orange5}{0.6800} & \colorcell{orange4}{0.5984} \\
\midrule
\multicolumn{8}{l}{\textit{Medical-specific VLMs}} \\
Lingshu-7B                 & \colorcell{orange4}{0.2305} & \colorcell{orange3}{0.6253} & \colorcell{orange4}{0.4582} & \colorcell{orange5}{0.6681} & \colorcell{orange5}{0.5755} & \colorcell{orange6}{0.7102} & \colorcell{orange5}{0.6112} \\
Lingshu-32B                & \colorcell{orange6}{0.2751} & \colorcell{orange6}{0.7120} & \colorcell{orange5}{0.4895} & \colorcell{orange6}{0.7011} & \colorcell{orange6}{0.5827} & \colorcell{orange6}{0.6991} & \colorcell{orange5}{0.6221} \\
Huatuo-GPT-Vision-7B       & \colorcell{orange1}{0.1901} & \colorcell{orange5}{0.6812} & \colorcell{orange6}{0.4997} & \colorcell{orange1}{0.5722} & \colorcell{orange2}{0.4990} & \colorcell{orange6}{0.6902} & \colorcell{orange3}{0.5850} \\
Huatuo-GPT-Vision-32B      & \colorcell{orange2}{0.2104} & \colorcell{orange6}{0.6995} & \colorcell{orange6}{0.5208} & \colorcell{orange3}{0.6153} & \colorcell{orange4}{0.5452} & \colorcell{orange6}{0.7204} & \colorcell{orange6}{0.6301} \\
\bottomrule[1.2pt]
\end{tabular}
}
\vspace{-10pt}
\end{table}

\subsection{Main Results Overview}
As shown in Table~\ref{tab:vlm_performance}, the overall performance comparison reveals several consistent trends across general-purpose and medical-specific VLMs. Scaling up general-purpose models generally improves performance, reflecting enhanced visual and linguistic capacity. However, this benefit begins to plateau at larger scales, suggesting diminishing returns in the absence of domain-specific adaptation. Meanwhile, medical-specific VLMs remain competitive, often matching or surpassing their larger general-purpose counterparts despite smaller model sizes. This highlights the importance of incorporating clinical knowledge during post-training.

Notably, no single model consistently outperforms others across all benchmarks, indicating that current VLMs lack universal generalisation across diverse medical tasks and modalities. These observations point toward the value of hybrid strategies that combine the broad generalisation of large-scale models with the precision of domain-adapted ones. Such approaches may offer a more scalable and effective path toward robust and clinically meaningful medical VLM.

\subsection{Understanding vs. Reasoning Performance}

To better characterise the strengths and limitations of current VLMs, we decompose their performance into understanding and reasoning components. As shown in Tables~\ref{tab:vlm_performance_retrieval} and~\ref{tab:vlm_performance_reasoning}, a consistent trend emerges: most models perform better on understanding tasks, which primarily involve visual recognition or factual retrieval. These tasks are generally more aligned with the models’ pretraining objectives and require less complex cross-modal inference.

In contrast, reasoning tasks continue to pose significant challenges for all models, irrespective of their size or medical focus. Such tasks typically require more complex reasoning and contextual integration, which current VLM still struggle to achieve. This gap underscores the need for targeted strategies such as reasoning-aware training objectives, modular architectures, or improved multimodal alignment to bridge the divide between perception and inference in real-world medical applications.

\section{Conclusion}
We present a comprehensive evaluation of general-purpose and medical-specific VLMs across a diverse set of medical benchmarks. By decomposing tasks into understanding and reasoning, we reveal consistent trends: model scaling improves performance but saturates, medical-specific tuning adds value, and reasoning remains a major bottleneck. Our findings highlight the need for hybrid strategies and more targeted evaluation to advance VLMs toward reliable clinical use.

%
%
%
\bibliographystyle{splncs04}
\bibliography{mybibliography}
\end{document}